%% file: main.tex
\documentclass{article}

\PassOptionsToPackage{numbers}{natbib}

    \usepackage[preprint]{neurips_2024}

\usepackage{graphicx} 
\usepackage{multirow} 
\usepackage{amsmath}
\usepackage[utf8]{inputenc} 
\usepackage[T1]{fontenc}    
\usepackage{hyperref}       
\usepackage{url}            
\usepackage{booktabs}       
\usepackage{amsfonts}       
\usepackage{nicefrac}       
\usepackage{microtype}      
\usepackage{xcolor}         
\usepackage{tabularx}
\usepackage{natbib}

\title{Pretrained Mobility Transformer: A Foundation Model for Human Mobility}

\author{%
  Xinhua Wu
  \hspace{1em}
  Haoyu He 
  \hspace{1em}
  Yanchao Wang 
  \hspace{1em}
  Qi Wang\\
  Department of Civil and Environmental Engineering\\
  Northeastern University\\
  \texttt{\{wu.xinh,he.haoyu1,wang.yanch,q.wang\}@northeastern.edu}
}

\begin{document}

\maketitle

\begin{abstract}
Ubiquitous mobile devices are generating vast amounts of location-based service data that reveal how individuals navigate and utilize urban spaces in detail. In this study, we utilize these extensive, unlabeled sequences of user trajectories to develop a foundation model for understanding urban space and human mobility. We introduce the \textbf{P}retrained \textbf{M}obility \textbf{T}ransformer (PMT), which leverages the transformer architecture to process user trajectories in an autoregressive manner, converting geographical areas into tokens and embedding spatial and temporal information within these representations.
Experiments conducted in three U.S. metropolitan areas over a two-month period demonstrate PMT's ability to capture underlying geographic and socio-demographic characteristics of regions. The proposed PMT excels across various downstream tasks, including next-location prediction, trajectory imputation, and trajectory generation. These results support PMT’s capability and effectiveness in decoding complex patterns of human mobility, offering new insights into urban spatial functionality and individual mobility preferences.
\end{abstract}

\section{Introduction}

Today, ubiquitous mobile devices are generating vast quantities of data, timestamped with both time and location, from large populations. This location-based service (LBS) data captures human mobility behaviors with unprecedented precision and sampling rates \cite{barbosa2018human}. Users' movement patterns across various urban areas provide comprehensive insight into individual behaviors related to work, leisure, entertainment, shopping, and more \cite{gonzalez2008understanding,wang2018urban}. The richness and detail of this data open up new avenues for understanding complex human behaviors in urban contexts. 

Despite the potential embedded in these vast resources, they are often underutilized due to the challenges of handling and interpreting high-volume, unstructured datasets. Particularly, the unlabeled sequences of user trajectories have not been fully explored and understood. The lack of research in these sequences means we missed the opportunity to gain a more in-depth understanding and knowledge of user movement, such as inter-area accessibility, area spatial functionality, and user mobility preferences, which can be extracted in an autoregressive manner and applied to downstream tasks.

To bridge both methodological and knowledge gaps, we propose \textbf{P}retrained \textbf{M}obility \textbf{T}ransformer (PMT) in this work. PMT offers a straightforward and viable framework for training a human mobility foundation model using LBS data. Specifically, each geographic area is abstracted as a specific token, with each token corresponding to a trainable spatial embedding. In this approach, user trajectories can be represented as a sequence composed of tokens, akin to tokens in large language models forming paragraphs. We trained the model using two types of autoregressive methods: predicting the next location and filling in masked locations. Considering the periodic nature of human mobility behaviors, we also designed temporal encoding to incorporate time information.

We conducted experiments within three metropolitan statistical areas (MSAs) in the United States, specifically Boston-Cambridge-Newton, MA-NH, Los Angeles-Long Beach-Anaheim, CA, and New York-Newark-Jersey City, NY-NJ-PA. The data spanned a two-month period, with the selected users representing 1\%-2\% of the total population. PMT consistently demonstrated exceptional performance across the different MSAs, accurately capturing local mobility patterns.
Remarkably, despite the absence of explicit latitude and longitude for geographic areas, our proposed PMT still constructed relative geographical relationships between different regions, as represented by the similarity of spatial embeddings. More interestingly, we found that the spatial embeddings of various regions also captured, to some extent, the socio-economic attributes of those areas, such as educational and income levels.

We highlight the contributions of this work as follows: \textbf{(1)} To the best of our knowledge, this is the first mobility foundation model trained extensively using large-scale LBS data. The proposed PMT, based on the transformer architecture, is a straightforward, effective, and scalable model for decoding complex patterns of human mobility. 
\textbf{(2)} The spatial embeddings in PMT effectively capture the geographical and socio-demographic characteristics of regions, providing a new perspective for understanding urban spaces with a data-driven approach. 
\textbf{(3)} The proposed PMT achieved superior performance consistently across multiple downstream tasks, including next-location prediction, trajectory imputation, and trajectory generation. PMT with a larger parameter size showed significant advantages in its performance.

\section{Related Work}

\textbf{Human Mobility Modelling.}
Statistical physicists have established theoretical models to uncover universal patterns in human mobility. CTRW \cite{brockmann2006scaling} describes human traveling behavior on various spatiotemporal scales by a two-parameter continuous-time random walk model, which ignores the considerable temporal and spatial regularity in individual movements \cite{gonzalez2008understanding, song2010modelling}. 
EPR \cite{pappalardo2015returners,pappalardo2016human} views the time duration between an individual's trips as a distribution and frames the process of choosing a travel destination as a probabilistic choice between exploring a new location and returning to a previously visited location. STS-EPR \cite{cornacchia2020modelling} further considers the impact of social factors during exploration and return behaviors. 
While these models offer valuable perspectives on human movement, their constrained parameters and predefined structures limit their capability in capturing individual's complex and heterogeneous mobility patterns.

\textbf{Mobility-related Tasks.} 
The development of deep learning and the availability of big mobility data have spurred a series of studies on tasks related to mobility. At the individual level, these tasks include next-location prediction \cite{feng2018deepmove,abideen2020deep,kong2018hst,dang2022predicting}, trajectory imputation \cite{chen2019complete,liu2021bidirectional,isufaj2023gti} and trajectory generation \cite{liu2018trajgans,wang2021large,he2023forecasting}. At the collective level, tasks include crowd flow prediction \cite{jiang2021deepcrowd,wang2020seqst,yao2019revisiting} and flow generation \cite{yao2020spatial,simini2020deep}. Notably, these proposed methodologies are tailored for specific types of tasks and leverage a diverse array of datasets, such as public bicycle data \cite{du2019deep}, taxi data \cite{sun2020predicting}, mobile phone record data \cite{barbosa2015effect,he2022percolation}, social media data \cite{cui2018social}, and points of interest (POI) check-in data \cite{gao2019predicting}. Some of these datasets, such as those from taxis and social media, represent small, biased samplings of the population. Additionally, given the variability in data availability and quality across different regions, reproducing these models in other areas poses significant challenges.
In contrast to these efforts, our work aims to develop a foundation model for understanding human mobility behavior, capable of directly solving or being fine-tuned to address various mobility-related tasks. To ensure the generalization of our approach, we exclusively utilize massive human mobility data generated by mobile devices for autoregressive training. The dataset has an original sampling rate of 8\%-10\% of the population.

\textbf{Geographic Embedding.}
The study of identifying the distinct functions (spatial semantics) of urban areas predominantly employed feature-based methodologies. This involves developing features for different regions based on information such as POI distribution and origin-destination flow \cite{gao2017extracting,wang2018using}. These features, depending on their interpretability and dimensions, can be further used in clustering algorithms to categorize functional areas \cite{yuan2012discovering}.
Recently, with the rise of representation learning \cite{bengio2013representation}, more studies captured the latent functions of areas with higher embedding dimensions using deep learning. In terms of the objects of embedding, past research has focused on embedding cellular towers \cite{song2021hermas}, road segments \cite{fu2020trembr,liu2017road2vec}, POIs \cite{yao2017sensing,yan2017itdl,zhai2019beyond}, geographic unit \cite{liu2020learning} and user trajectories \cite{fu2020trembr,gao2017identifying}. From the perspective of embedding methods, \cite{liu2017road2vec,yao2017sensing,yan2017itdl,zhai2019beyond} utilized the skip-gram model proposed by word2vec \cite{mikolov2013efficient}, \cite{liu2020learning} applied graph neural networks with the idea of node2vec \cite{grover2016node2vec}, and \cite{song2021hermas,fu2020trembr,gao2017identifying} incorporated  encoder-decoder RNN structures. Our work distinguishes itself in two respects. First, in terms of the embedding objects, we directly embed urban areas (polygons) rather than just parts of the city like POIs, aiming for a more comprehensive understanding of human mobility behavior. This is similar to \cite{liu2020learning}, but notably we use detailed individual mobility records rather than aggregated origin-destination flow for training. Second, methodologically, we employ large pretrained transformer models to extract embeddings of geographic areas, which is motivated by the breakthroughs of pretraining models in NLP, such as BERT \cite{devlin2018bert} and GPT \cite{brown2020language}.

\section{Preliminary}
\label{sec:preliminary}

\begin{figure}[!htbp]
  \centering
  \includegraphics[width=2.5in]{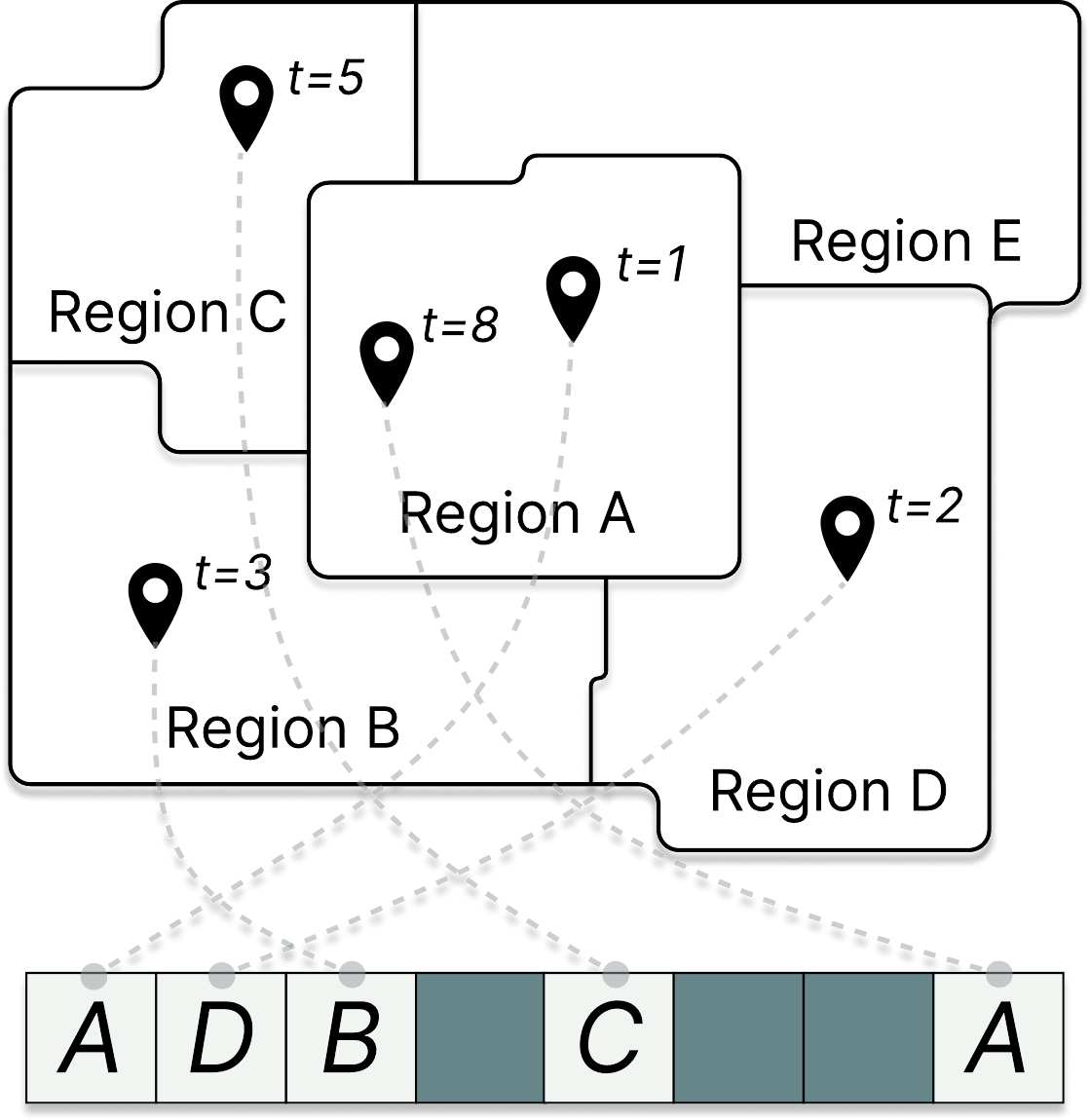}
  \caption{An illustrative user trajectory sequence with \(T=8\) time steps. Location information is available for five time steps, denoted by light-colored boxes. The remaining three time steps, which lack user location data, are represented by dark-colored boxes.}
  \label{fig:preliminary}
\end{figure}

\textbf{Location-based Service Data.} The widespread use of mobile devices with GPS has led to the creation of large amounts of data from location-based services. Each entry in LBS data contains the geographical coordinates, specifically the latitude and longitude of the user, as well as the timestamp in which the data was recorded. A snapshot of synthetic LBS data is provided in Appendix A.
Despite the high volumes and spatiotemporal scales, LBS data faces some limitations, including spatial inaccuracies and inconsistent sampling time and rates. To address these issues, spatial and temporal aggregation is typically performed, as shown in Figure \ref{fig:preliminary}. Each user is assigned to a spatial region based on their locations within a specific time period. Due to the incompleteness of LBS data, the user location could be unknown within some time windows, which is represented by the dark boxes in Figure \ref{fig:preliminary}.

\textbf{User Trajectory Sequence.} In this study, we model a user's movement using a sequence of location identifiers over consecutive time intervals, represented as: \(X = (x_1, x_2, x_3, \ldots, x_t, \ldots, x_T)\). Here, $T$ denotes the total number of time windows, and each $x_t$ is an identifier for a U.S. Census Block Group (CBG) where the user is located at time \(t\). CBGs are geographical units used in the U.S. Census, typically containing between 600 and 3,000 residents. We use CBGs to track locations because they offer a good balance between detail and privacy; they are small enough to provide significant location data yet large enough to prevent precise tracking of individual users (e.g., the exact locations of homes and workplaces). A placeholder token is used if the location data is missing. Each time window is set to 30 minutes. The time window duration aligns with the frequency of location data collection, which averages 50 to 100 location records per user per day. Choosing a 30-minute window helps to avoid overly sparse sequences. In cases where a user's data shows movement across different CBGs within the same time window, we only record the last CBG in which the user was located.
\textit{Temporal occupancy} is defined as the proportion of elements in the sequence that have specific geographic locations. For example, the sequence shown in Figure \ref{fig:preliminary} has a \textit{temporal occupancy} of \( \nicefrac{5}{8} = 62.5\% \) .

\section{Methods}

\begin{figure}[!htbp]
  \centering
  \includegraphics[width=\textwidth,keepaspectratio]{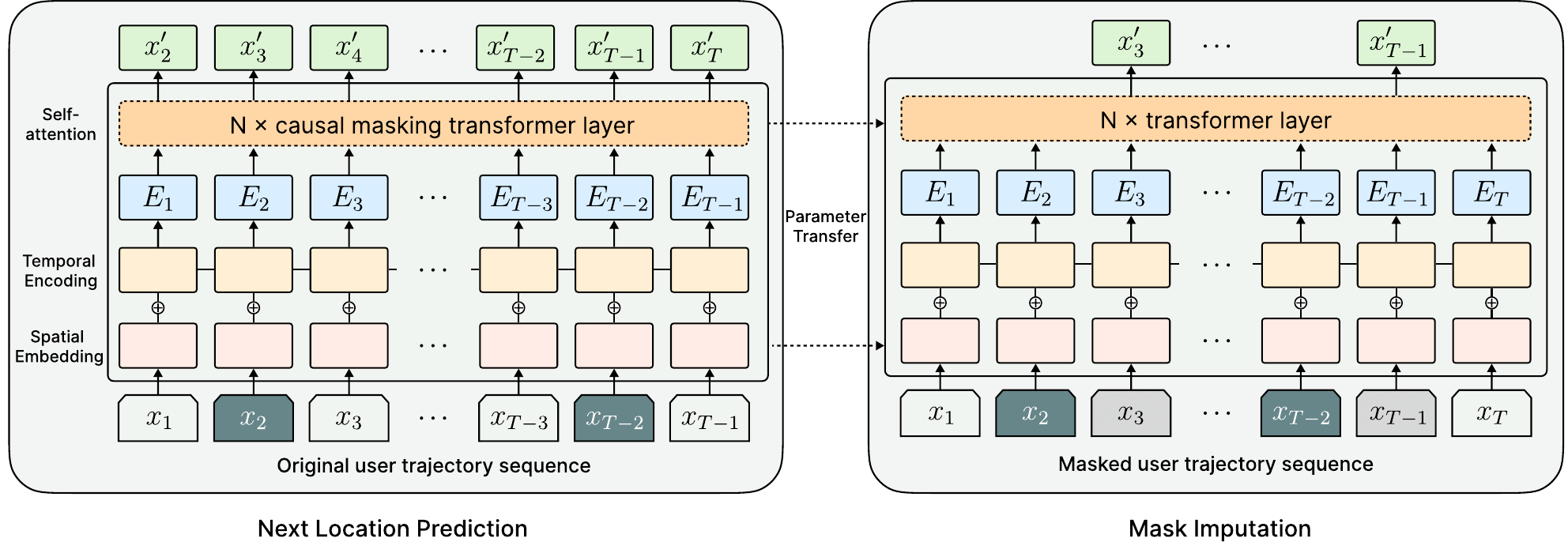}
  \caption{Overall pretraining procedures for PMT. In the input layer, light-colored, dark-colored, and gray-shadowed boxes distinctly represent elements in the trajectory sequence that possess location information, are devoid of location information, and have obscured location information, respectively. Apart from the causal masking, the same architectures are used in both pretraining tasks.}
  \label{fig:model}
\end{figure}

We introduce PMT and its implementation in this section. PMT’s model architecture is a multi-layer transformer based on the original implementation of \cite{vaswani2017attention} without the encoder-decoder structure. The overall framework of per-training is shown in Figure \ref{fig:model}. The trajectory sequences outlined in Section \ref{sec:preliminary} serve as inputs for autoregressive training and the embedded sequence is processed by multiple layers of transformer blocks to generate the sequences required for a specific pretraining task.

In this work, we employ three models with parameter sizes of 1.6M, 21M, and 550M, respectively. These models are characterized by different dimensions: embedding size (\(D)\), hidden size (\(H)\), the number of transformer layers (\(L)\), and the number of self-attention heads (\(A\)). The specifications for these models are as follows: PMT-1.6M (\(D=128, H=256, L=4, A=4\)), PMT-21M (\(D=512, H=1024, L=8, A=8\)), and PMT-550M (\(D=1024, H=2048, L=16, A=16\)).

\subsection{Spatiotemporal Embedding}
Compared to pretraining models for NLP \cite{brown2020language} and CV \cite{kirillov2023segment}, the interaction between geographical space and time in trajectory sequences is more crucial and complex. For example, the Central Business District (CBD) during the day and residential areas at night are hotspots for human activities. Additionally, there is a distinct periodicity (e.g., 24 hours, 7 days, etc.) in individual travel behaviors, reflecting consistent patterns over time.
To better capture the spatiotemporal relationships, we divided the sequence embedding into spatial embedding and temporal encoding, with their sum constituting the final embedding. Spatial embedding maps each predefined spatial area (i.e., CBG) to a fixed-length vector, similar to how tokens are mapped to vectors in large language models. The parameters for spatial embedding are learnable and initialized randomly without any prior knowledge.

In terms of temporal encoding, we employed a predefined encoding scheme instead of using learnable parameters. 
In this work, temporal encoding includes absolute time encoding and periodic encoding. Absolute time encoding, identical to positional encoding in \cite{vaswani2017attention}, is designed to furnish comprehensive temporal information within trajectory sequences. Meanwhile, periodic encoding injects crucial periodical information that reflects human mobility patterns, specifically daily encoding and weekly encoding. Periodic encoding facilitates more efficient model pretraining and enhances the model's ability to generalize across different temporal contexts in downstream tasks.

Let \(X\) be a trajectory sequence of length \(T\), and \(D\) the desired embedding dimension. Temporal Encoding, \( TE \in \mathbb{R}^{T \times D} \), is the concatenation of absolute time encoding, \( \text{AE} \in \mathbb{R}^{T \times D/2}\), daily encoding, \( \text{DE} \in \mathbb{R}^{T \times D/4}\), and weekly encoding, \( \text{WE} \in \mathbb{R}^{T \times D/4}\). \text{AE}, \text{DE} and \text{WE} are defined as Equation \ref{eq:te}. A visualization of the implemented temporal encoding is presented in Appendix B.

\begin{equation}
\label{eq:te}
\begin{aligned}
\text{AE}_{(t, j)} &= 
\begin{cases} 
\sin\left(\frac{t}{10000^{4i/D}}\right) & \text{if } j = 2i \\
\cos\left(\frac{t}{10000^{4i/D}}\right) & \text{if } j = 2i+1 
\end{cases} \\
\text{DE}_{(t, j)} &= 
\begin{cases} 
\sin\left(\frac{\theta_d(t)}{10000^{8i/D}}\right) & \text{if } j = 2i \\
\cos\left(\frac{\theta_d(t)}{10000^{8i/D}}\right) & \text{if } j = 2i+1 
\end{cases} \\
\text{WE}_{(t, j)} &= 
\begin{cases} 
\sin\left(\frac{\theta_w(t)}{10000^{8i/D}}\right) & \text{if } j = 2i \\
\cos\left(\frac{\theta_w(t)}{10000^{8i/D}}\right) & \text{if } j = 2i+1 
\end{cases}
\end{aligned}
\end{equation}

where \(t\) is the index of the time step and \(j\) is the dimension.  \(\theta_d(t)\) and \(\theta_w(t)\) are the angular coordinates for daily and weekly cycles, respectively, both mapped to the interval \([0, 2\pi]\). Given the 30-minute time window, \(\theta_d(t + 48i) = \theta_d(t)\) and \(\theta_w(t + 7\times 48i) = \theta_w(t)\) for all integer \(i\).

\subsection{Pretraining Tasks}

Two different pretraining tasks, next-location prediction and mask imputation, are designed for the training of PMT. These pretraining tasks are analogous to the pretraining methods of GPT \cite{radford2018improving} and BERT \cite{devlin2018bert}, respectively.

\textbf{Next-location Prediction Task.} As shown on the left side of Figure \ref{fig:model}, with the user trajectories as input, the model is to predict the next location given the previous trajectory. A causal masking strategy is applied to prevent the leakage of future information. As mentioned above, the dark-colored boxes in Figure \ref{fig:model} indicate time steps where location data is missing in the user's trajectory. We hypothesize that these instances of missing data might subtly reflect the user's movement patterns. For instance, when a user reaches their destination, they are likely to turn off the navigation app, ceasing to upload their location. 
Consequently, sequence positions lacking geographical information are processed as a special token rather than omitted or masked. They share a learnable spatial embedding and are utilized in subsequent self-attention transformer layers; however, due to the absence of ground truth, these positions are not included in the loss calculation.

\textbf{Mask Imputation Task.} In this task, we randomly mask positions within a trajectory sequence, denoted by the gray-shadowed boxes on the right side of Figure \ref{fig:model}, and task the model with predicting these obscured geographic locations. Similarly, masks where the ground truth is missing are not used to calculate the loss. Causal masks were not used, allowing the model to complete the sequence from all unmasked positions.
We set the masking probability at 50\% to create a challenging scenario that simple interpolation would not be able to accomplish. However, this high level of masking, combined with pre-existing missingness in geographical information within the sequence, considerably destabilized the training process, especially for the PMT-550M.
To address this issue, we initialized the model with available parameters from the Next-location Prediction Task. Here, a reasonable suspicion is whether this could lead to leakage of future trajectory. Nonetheless, our subsequent experiments (see Appendix E) showed consistent imputation performance across both training and testing datasets, indicating that the model is not affected by any significant data leakage during pretraining.

\section{Experiments and Results}

\subsection{Experiment Setting}
In this study, mobile location data from the three metropolitan statistical areas (MSAs) introduced above were used for experiments. Hereafter, they are referred to as BOS, LA, and NYC, respectively. Data in these three MSAs are described in Table \ref{tab:data}.

The data spans from January 1, 2020, to February 28, 2020, making the length of each trajectory sequence 2,832. We selected sequences with a \textit{temporal occupancy} (see Section \ref{sec:preliminary}) greater than 50\%.  The sequences were allocated into two groups, with two-thirds designated for pretraining and the remaining one-third for testing the model's performance on downstream tasks. This allocation was to guarantee that there was no overlap of individuals between the pretraining and the testing. More details regarding training are presented in Appendix C.

\begin{table}[!htbp]
  \caption{Data description for three MSAs. Temporal occupancy in the table refers to the average temporal occupancy across all trajectory sequences, while the sampling rate indicates the ratio of the number of filtered trajectory sequences to the total population.}
  \label{tab:data}
  \centering
  \begin{tabular}{lccccc}
    \toprule
    MSA   & Population & CBG & Sequence & Average temporal occupancy & Sampling rate \\
    \midrule
    BOS & 4.9M & 3.5k & 59k & 75.4\% & 1.20\%     \\
    LA & 12.9M & 8.2k & 249k & 77.2\% & 1.93\%     \\
    NYC  & 23.6M & 14.4k & 465k & 76.5\% & 1.97\%     \\
    \bottomrule
  \end{tabular}
\end{table}

\subsection{What does the Spatial Embedding Say?}

In this subsection, we explore the meaning of spatial embeddings in PMT that represent geographic areas (i.e., CBGs). Despite PMT models being trained solely on unlabeled trajectory sequences, we discovered that these spatial embeddings effectively capture the geographical and deeper socio-demographic characteristics of the CBGs.

\textbf{Spatial Embedding Similarity and Geographic Distance.} We calculated the cosine similarity (normalized to 0-1) among all CBGs within the same MSA and the distances (Euclidean distance) between the centroids of all CBGs. Table \ref{tab:similarity-distance} shows their negative correlation, indicating that CBGs with higher similarity are also geographically closer to each other. 
PMT-550M and PMM-21M models exhibit a weaker negative correlation, potentially attributable to two factors. First, these models utilize deeper transformer architectures, which may produce more abstract representations in their embedding layers. Second, these models likely construct spatial embeddings that go beyond geographic proximity. For example, factors such as transportation facilities and public transport can also impact the accessibility of CBGs.

\begin{table}[!htbp]
  \caption{The correlation between spatial embedding similarity and Euclidean distance.}
  \label{tab:similarity-distance}
  \centering
  \begin{tabularx}{\textwidth}{l*{6}{>{\centering\arraybackslash}X}}
    \toprule
    \multirow{2}{*}{Model} & \multicolumn{3}{c}{Pearson correlation coefficient} & \multicolumn{3}{c}{Spearman's correlation coefficient} \\
    \cmidrule(lr){2-4} \cmidrule(lr){5-7}
    & BOS & LA & NYC & BOS & LA & NYC \\
    \midrule
    PMT-1.6M & -0.29 & -0.18 & -0.34 & -0.34 & -0.19 & -0.38 \\
    PMT-21M  & -0.11 & -0.17 & -0.15 & -0.15 & -0.21 & -0.19 \\
    PMT-550M & -0.06 & -0.12 & -0.07 & -0.09 & -0.14 & -0.10 \\
    \bottomrule
  \end{tabularx}
\end{table}

\begin{figure}[!htbp]
  \centering
  \includegraphics[width=\textwidth,keepaspectratio]{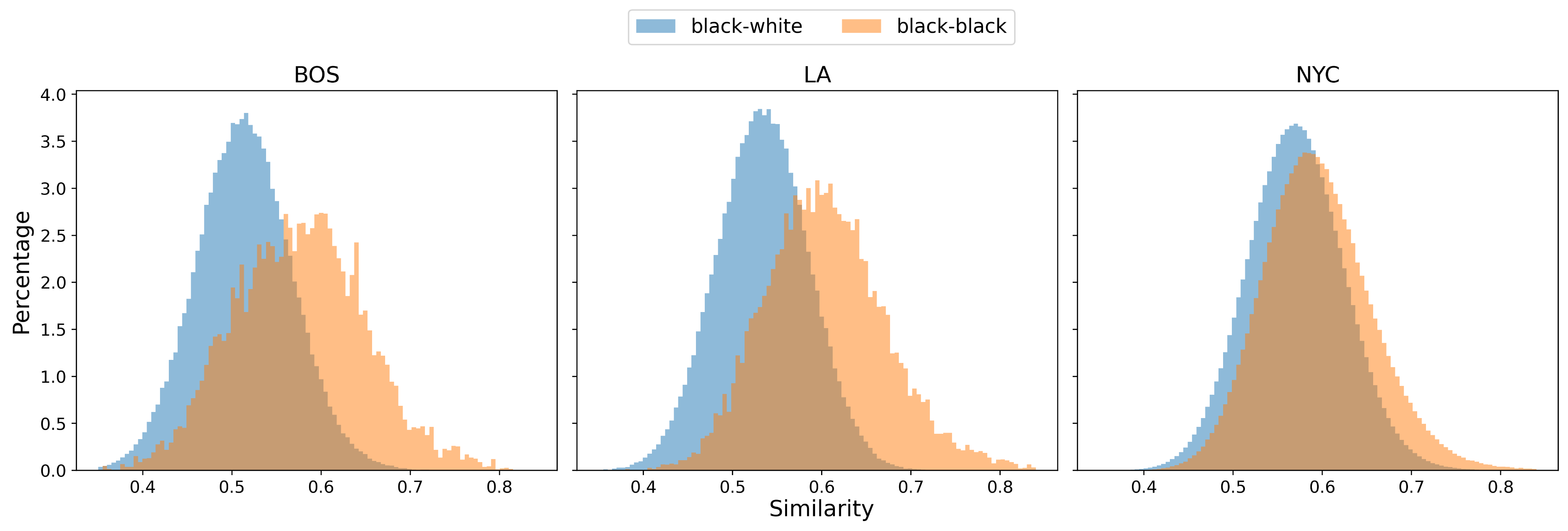}
  \caption{Spatial embedding similarity distribution of black-black comparison and black-white comparison (PMT-1.6M).}
  \label{fig:similarity-black}
\end{figure}

\textbf{Spatial Embedding Similarity and Racially Dominant CBGs.}
Using the data from the American Community Survey, we define CBGs where over 50\% of the population is black as black-dominant CBGs, and those where over 50\% of the population is White as white-dominant CBGs. The black-black comparison calculates the similarity between all pairs of CBGs within black CBGs, while the black-white comparison calculates the similarity of all CBG pairs composed of one black CBG and one white CBG.
Figure \ref{fig:similarity-black} shows that the distribution of black-black similarity is significantly shifted to the right compared to the black-white similarity across three MSAs, suggesting a stronger cohesion within black-dominant CBGs. This could also indicate that minority populations are more likely to participate in activities within their own racial communities, beyond simply living in close proximity.

\textbf{Spatial Embedding Similarity and Socio-demographics.} Individuals from different socio-economic backgrounds often display distinct behavioral patterns. We theorize that spatial embeddings in PMT, generated through autoregressive training using extensive user trajectory data, might contain socio-demographic information specific to certain areas. To validate this hypothesis, we employed CBGs' spatial embedding as inputs to predict their socio-demographic characteristics. To control for randomness, we conducted one hundred trials, randomly selecting half of the CBGs for training and the other half for validation in each trial. A simple neural network with two hidden layers was trained over five epochs per trial. The results, as shown in Table \ref{tab:embed_pred}, indicate that this model can predict the population, education levels, and income with moderate accuracy, suggesting that PMT embeddings indeed incorporate socio-demographic information, indicating that PMT may implicitly model urban characteristics and functionalities during its pre-training.

\begin{table}[!htbp]
  \caption{$R^2$ of demographic prediction using spatial embedding (PMT-1.6M).}
  \label{tab:embed_pred}
  \centering
  \begin{tabular}{lccc}
    \toprule
       & BOS & LA & NYC \\
    \midrule
    Resident population & \(0.296 \pm 0.016\) & \(0.199 \pm 0.010\) & \(0.185 \pm 0.009\)     \\
    Bachelor's degree proportion  &  \(0.475 \pm 0.014\) & \(0.558 \pm 0.011\) & \(0.483 \pm 0.010\)    \\
    Median household income  & \(0.321 \pm 0.012\) & \(0.380 \pm 0.011\) & \(0.409 \pm 0.009\)  \\
    \bottomrule
  \end{tabular}
\end{table}

\subsection{Downstream Task Evaluations}
We evaluated the performance of the PMT on several downstream tasks, including next-location prediction, trajectory imputation, and trajectory generation. We found that the proposed PMT models (PMT-21M and PMT-550M) significantly outperformed baselines (see Appendix D) across various tasks. Even the PMT-1.6M, with significantly fewer parameters than the baselines, exhibited superior performance in some tasks. Additionally, there seems to be a scaling law; increasing the model parameters benefits PMT in encoding spatiotemporal information and human mobility, thereby enhancing performance across various downstream tasks.

\textbf{Next Location Prediction.} 
This task, mirroring our first pretraining one, is to predict the next location of a user, given their previous trajectory. The task is critical for accurate next-location prediction is crucial for personalized recommendation services, such as POI recommendations. The historical trajectory length ranges from 30 minutes to two months to simulate cold-start and warm-start scenarios in next-location prediction.

Table \ref{tab:prediction} shows the prediction results for both daytime and the entire day separately. Given the highly predictable nature of human behavior during the night, daytime predictions exhibit lower accuracy compared to predictions for the entire day. Notably, PMT-550M and PMT-21M consistently outperform baselines across all MSAs. While PMT-1.6M performs better than the baselines in BOS, its performance is inferior to the baselines in LA and NYC. This could be attributed to the higher number of CBGs in these two MSAs, where PMT-1.6M might not adequately capture the rich and complex patterns with its small parameter size.

\begin{table}[!htbp]
  \caption{Performance Metrics for Next-Location Prediction. This table presents the cross entropy loss (without label smoothing) for predictions made during daytime hours, defined as 7:30 AM to 7:30 PM local time. The best-performing result is highlighted in bold, while the second-best is underlined for easy comparison.}
  \label{tab:prediction}
  \fontsize{9}{11}\selectfont
  \centering
  \begin{tabular}{llcccccccc}
    \toprule
     & \multirow{2}{*}{Model} & \multicolumn{4}{c}{Daytime} & \multicolumn{4}{c}{All Day} \\
    \cmidrule(r){3-6} \cmidrule(r){7-10}
    & & Loss & Acc@1 & Acc@3 & Acc@10 & Loss & Acc@1 & Acc@3 & Acc@10 \\
    \midrule
    \multirow{6}{*}{\rotatebox[origin=c]{90}{BOS}} 
    & FBM         & \textit{/} & 57.72\% & 82.96\% & 90.87\% & \textit{/} & 70.87\% & 88.13\% & 93.63\% \\
    & LSTM        & 1.31 & 79.07\% & 85.93\% & 90.23\% & 0.82 & 86.49\% & 91.08\% & 93.80\% \\
    & DeepMove    & 1.11 & 80.13\% & 87.44\% & 91.96\% & 0.71 & 87.15\% & 92.14\% & 94.97\% \\
    & PMT-1.6M    & 1.15 & 80.45\% & 88.14\% & 92.40\% & 0.78 & 87.37\% & 92.58\% & 95.24\%\\
    & PMT-21M     & \underline{1.05} & \underline{81.71\%} & \underline{89.01\%} & \underline{93.23\%} & \underline{0.70} & \underline{88.23\%} & \underline{93.16\%} & \underline{95.79\%} \\
    & PMT-550M    & \textbf{1.02} & \textbf{81.81\%} & \textbf{89.23\%} & \textbf{93.58\%} & \textbf{0.67} & \textbf{88.28\%} & \textbf{93.29\%} & \textbf{96.01\%} \\
    \midrule
    \multirow{6}{*}{\rotatebox[origin=c]{90}{LA}} 
    & FBM         & \textit{/} & 57.19\% & 81.04\% & 88.44\% & \textit{/} & 69.36\% & 86.40\% & 91.64\% \\
    & LSTM        & 1.38 & 78.93\% & 85.21\% & 88.91\% & 0.92 & 85.98\% & 90.38\% & 92.78\% \\
    & DeepMove    & 1.22 & 79.73\% & 86.19\% & 90.18\% & 0.81 & 86.54\% & 91.11\% & 93.67\% \\
    & PMT-1.6M    & 1.37 & 79.49\% & 86.23\% & 90.05\% & 0.98 & 86.19\% & 90.96\% & 93.47\% \\
    & PMT-21M     & \underline{1.13} & \textbf{81.70\%} & \underline{87.92\%} & \underline{91.72\%} & \underline{0.78} & \textbf{87.84\%} & \underline{92.17\%} & \underline{94.60\%} \\
    & PMT-550M    & \textbf{1.11} & \underline{81.64\%} & \textbf{88.10\%} & \textbf{92.00\%} & \textbf{0.76} & \underline{87.76\%} & \textbf{92.29\%} & \textbf{94.79\%} \\
    \midrule
    \multirow{6}{*}{\rotatebox[origin=c]{90}{NYC}} 
    & FBM         & \textit{/} & 56.86\% & 82.05\% & 89.51\% & \textit{/} & 69.30\% & 86.76\% & 92.19\% \\
    & LSTM        & 1.83 & 76.85\% & 81.64\% & 85.17\% & 1.24 & 84.36\% & 87.68\% & 90.07\% \\
    & DeepMove    & 1.18 & 79.75\% & 86.54\% & 90.78\% & 0.80 & 86.34\% & 91.17\% & 93.93\% \\
    & PMT-1.6M    & 1.48 & 78.13\% & 85.37\% & 90.00\% & 1.15 & 84.35\% & 89.77\% & 93.11\% \\
    & PMT-21M     & \underline{1.08} & \underline{81.98\%} & \underline{88.39\%} & \underline{92.42\%} & \underline{0.76} & \underline{87.80\%} & \underline{92.32\%} & \underline{94.45\%} \\
    & PMT-550M    & \textbf{1.04} & \textbf{82.02\%} & \textbf{88.71\%} & \textbf{92.80\%} & \textbf{0.72} & \textbf{87.84\%} & \textbf{92.55\%} & \textbf{95.22\%}\\
    \bottomrule
  \end{tabular}
\end{table}

\textbf{Trajectory Imputation.} 
As users have control over uploading their location information, trajectory sequences derived from LBS data frequently exhibit low temporal occupancy. This constraint hampers a comprehensive understanding of individual mobility patterns. The aim of trajectory imputation is to reconstruct user locations for all time intervals based on existing incomplete sequences. We specifically assessed the model's ability to fill in missing data across various levels of data sparsity. We randomly removed 50\% to 90\% of the location information from the trajectory sequences over a two-month period. This simulates different degrees of sparsity observed in location-based service data.

\begin{figure}[!htbp]
  \centering
  \includegraphics[width=\textwidth,keepaspectratio]{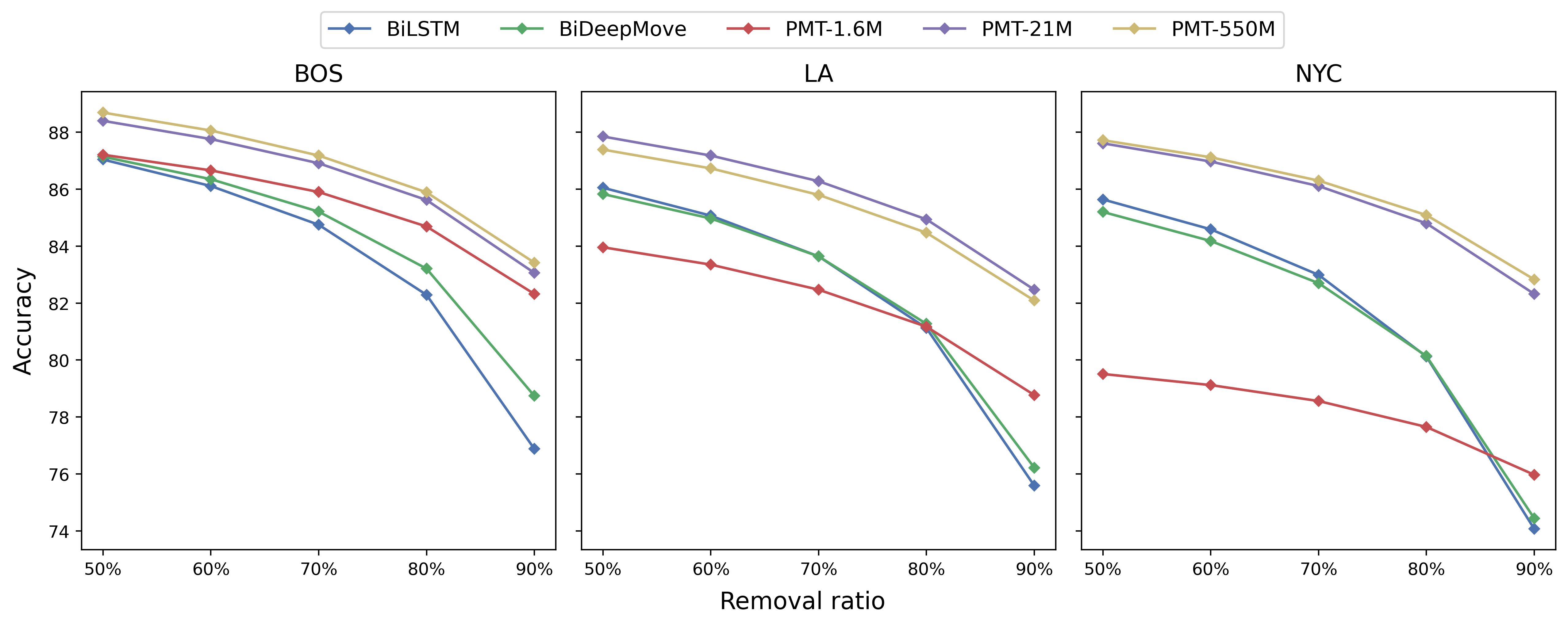}
  \caption{Accuracy of imputation across varying sparsity levels. The removal ratio represents the proportion of locations randomly omitted from the trajectory sequences. Considering the pre-existing gaps in the sequences, the actual missingness rate is higher than indicated by the removal ratio alone. At a 90\% removal ratio, the average temporal occupancy of sequences is approximately 7.5\%.}
  \label{fig:imputation}
\end{figure}

Figure \ref{fig:imputation} shows the model's imputation accuracy across different degrees of data sparsity. PMT-550M and PMT-21M consistently outperform other models across a range of sparsity levels, whereas PMT-1.6M, despite initially lagging behind baselines at lower missingness rates, surpasses them as sparsity increases. This suggests PMT's advantages in capturing the spatiotemporal regularities of human mobility from sparse trajectories. PMT-550M and PMT-21M exhibit comparable accuracy, but PMT-550M shows better performance in terms of cross-entropy, as detailed in Appendix E.

\textbf{Trajectory Generation.} Differing from next-location prediction, trajectory generation aims to produce extended sequences of human mobility behaviors over a longer duration. This can inform potential future traffic flows and offer reasonable synthetic trajectories for mobility simulation. The long-term generation is achieved through multi-step predictions by the model. Maximum probability-based sampling was employed for both baselines and PMT.

Table \ref{tab:generation} presents the results of 8-hour (9 am-5 pm) and 24-hour (9 am-9 am) trajectory generation. We utilized a modified n-gram precision \cite{papineni2002bleu} (see Appendix D) at the individual level to assess the similarity between generated trajectories and actual trajectories, where a higher n-gram precision indicates better representation of user stay and movement behaviors across different areas. Additionally, we aggregated the generated and actual trajectories, computing discrepancies in the number of individuals locating in different CBGs. A smaller aggregated level error indicates a better representation of human macroscopic distribution across the city.
An interesting observation is that, despite PMT-1.6M performing well at the individual level, it fails to accurately replicate the aggregated level population distribution. However, with an increase in model parameters, PMT-550M and PMT-21M significantly outperform the baselines both at the individual and aggregated levels. This suggests that larger PMT models can more effectively capture the temporal characteristics of human mobility.

\begin{table}
  \caption{Performance of generation. A modified n-gram precision score is used for individual-level evaluation and MAPE is calculated only for the most populous 25\% of CBGs at each time interval. The best result is bolded, and the second best is underlined in the table.}
  \label{tab:generation}
  \fontsize{9.5}{11}\selectfont
  \centering
  \begin{tabular}{llcccccccc}
    \toprule
     & \multirow{3}{*}{Model} & \multicolumn{4}{c}{Individual level} &  \multicolumn{4}{c}{Aggregated level} \\
    \cmidrule(r){3-6} \cmidrule(r){7-10} 
    & &  \multicolumn{2}{c}{8 hour} & \multicolumn{2}{c}{24 hour} & \multicolumn{2}{c}{8 hour} & \multicolumn{2}{c}{24 hour} \\
    \cmidrule(r){3-4} \cmidrule(r){5-6} \cmidrule(r){7-8} \cmidrule(r){9-10}
    & & 2-gram & 3-gram & 2-gram & 3-gram & RMSE & MAPE & RMSE & MAPE \\
    \midrule
    \multirow{5}{*}{\rotatebox[origin=c]{90}{BOS}} 
    & LSTM        & 0.532 & 0.481 & 0.448 & 0.415 & 2.72 & 29.87\% & 4.18 & 45.78\% \\
    & DeepMove    & 0.538 & 0.479 & 0.622 & 0.582 & 2.27 & 27.39\% & 3.21 & 30.93\%  \\
    & PMT-1.6M   & 0.634 & 0.570 & 0.765 & 0.720 & 2.91 & 26.80\% & 5.11 & 20.66\% \\
    & PMT-21M     & \underline{0.642} & \underline{0.578} & \textbf{0.766} & \textbf{0.721} & \underline{2.21} & \textbf{24.52\%} & \underline{2.36} & \textbf{18.40\%}  \\
    & PMT-550M   & \textbf{0.646} & \textbf{0.581} & \underline{0.763} & \underline{0.718} & \textbf{2.10} & \underline{25.29\%} & \textbf{2.10} & \underline{19.73\%}  \\
    \midrule
    \multirow{5}{*}{\rotatebox[origin=c]{90}{LA}} 
    & LSTM        & 0.528 & 0.478 & 0.479 & 0.442 & 4.52 & 25.07\% & 7.72 & 38.62\% \\
    & DeepMove    & 0.545 & 0.494 & 0.622 & 0.582 & 3.89 & 23.30\% & 5.25 & 26.28\% \\
    & PMT-1.6M   & 0.636 & 0.576 & 0.754 & 0.707 & 4.36 & 22.43\% & 8.02 & 20.66\% \\
    & PMT-21M    & \textbf{0.657} & \textbf{0.594} & \underline{0.764} & \underline{0.715} & \underline{3.37} & \textbf{20.71\%} & \underline{3.40} & \underline{15.61\%}  \\
    & PMT-550M   & \underline{0.651} & \underline{0.589} & \textbf{0.764} & \textbf{0.715} & \textbf{3.19} & \underline{20.85\%} & \textbf{2.81} & \textbf{15.41\%}  \\
    \midrule
    \multirow{5}{*}{\rotatebox[origin=c]{90}{NYC}} 
    & LSTM        & 0.489 & 0.442 & 0.466 & 0.431 & 5.98 & 28.89\% & 7.96 & 42.01\% \\
    & DeepMove       & 0.558 & 0.507 & 0.583 & 0.545 & 3.89 & 21.63\% & 6.79 & 29.70\% \\
    & PMT-1.6M   & 0.616 & 0.549 & 0.736 & 0.690 & 6.66 & 24.11\% & 13.95 & 19.56\% \\
    & PMT-21M    & \textbf{0.659} & \textbf{0.586} & \textbf{0.772} & \textbf{0.724} & \underline{3.36} & \underline{20.26\%} & \underline{3.27} & \textbf{14.75\%} \\
    & PMT-550M   & \underline{0.653} & \underline{0.581} & \underline{0.768} & \underline{0.721} & \textbf{3.31} & \textbf{20.16\%} & \textbf{3.23} & \underline{14.82\%} \\
    \bottomrule
  \end{tabular}
\end{table}

\section{Discussion}

This study introduces the PMT, a model designed to decode intricate patterns of human mobility using extensive LBS data. Through experiments conducted in three U.S. metropolitan areas, we found that PMT is able to not only predict and complete trajectory sequences accurately but also capture geographic and socio-demographic attributes. PMT excels across various downstream tasks, underscoring its effectiveness in modeling human mobility and urban spaces. This work represents a significant advancement in leveraging data-driven approaches to unravel the complexities of urban spatial functionality and individual mobility preferences, paving the way for further insights and explorations in urban planning, transportation, mobility, and related domains.

Despite the contribution, two concerns remain regarding the human mobility foundation model. The first one is the issue of sample imbalance. Although the sampling rate of LBS data is relatively high, it originates from mobile devices. Elderly people and the extremely impoverished who do not use smart devices may be overlooked in the data. Policies informed by the model could be unfair to these populations. The second one is the concern about user privacy. Even though we control the trajectory resolution at the CBG level, we are concerned that by fine-tuning with known personal information of a few users, PMT could accurately infer private details such as gender, age, race, and income based on users' long-term trajectories.

\bibliographystyle{unsrt}
\bibliography{references}

\newpage
\include{supplementary}



\end{document}

%% file: supplementary.tex







\appendix

\section{Location-based service data}

\begin{figure}[!htbp]
  \centering
  \includegraphics[width=3.5in,keepaspectratio]{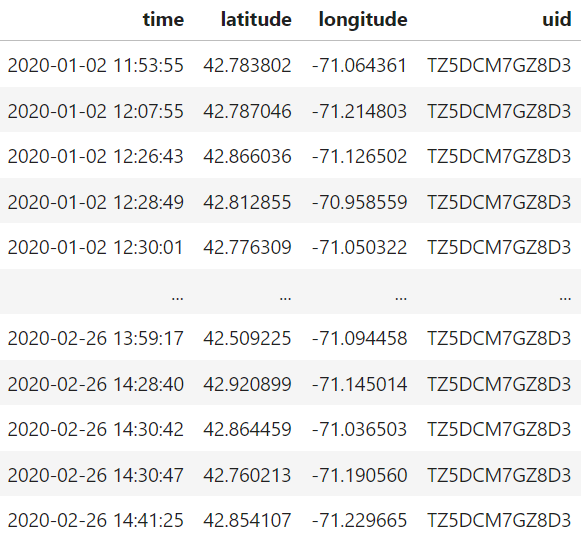}
  \caption{A synthetic example of LBS data. It is important to note that, in order to protect user privacy, the latitude and longitude of a user's primary addresses (such as home or workplace) will be shifted to the centroid of the corresponding CBG.}
\end{figure}

\section{Temporal Encoding}

\begin{figure}[!htbp]
  \centering
  \includegraphics[width=\textwidth,keepaspectratio]{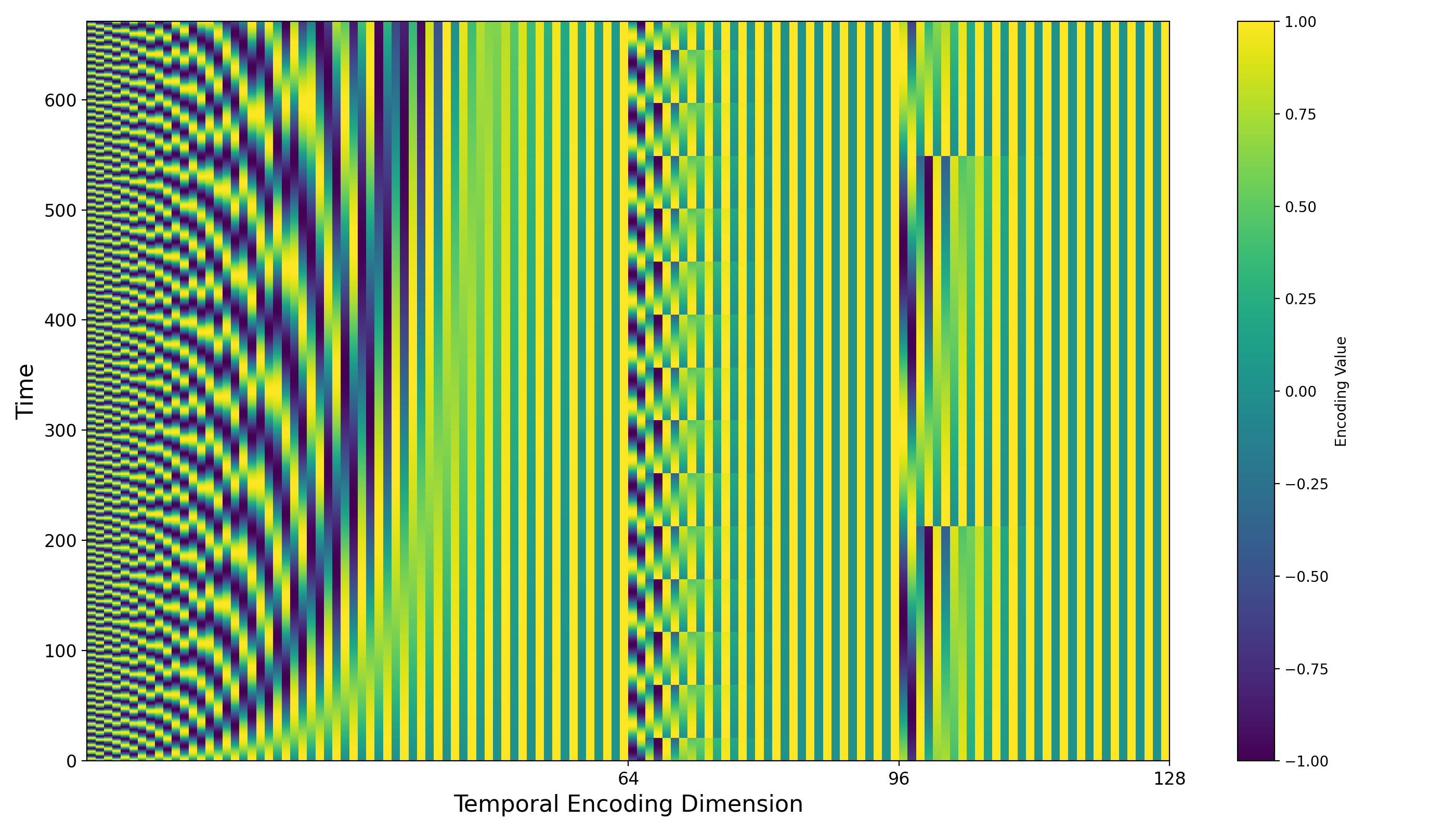}
  \caption{A temporal encoding of 256 dimensions. The leftmost 128 dimensions represent absolute time encoding, dimensions from 129 to 192 in the middle represent day encoding, and the rightmost 64 dimensions represent week encoding.}
\end{figure}

\section{Training Details}

We trained our models on one machine with 4 NVIDIA A100 @80G GPUs. In our PMT-550M architecture, each training step required approximately 3.5 seconds for a batch size of 16 and 6.0 seconds for a batch size of 32. To mitigate overfitting, we employed label smoothing \cite{muller2019does} and dropout \cite{srivastava2014dropout} regularization techniques. The Adam optimizer \cite{kingma2014adam} was used with the following parameters: $\beta_1 = 0.9$, $\beta_2 = 0.99$, and $\epsilon = 10^{-9}$. A dynamic learning rate similar to \cite{vaswani2017attention} is used over the training, according to the following formula, with the maximum upper bound of the learning rate set to 1e-4:

\[
\text{lr} = \frac{1}{\sqrt{d_{\text{Embedding}}}} \times \min\left(\frac{1}{\sqrt{\text{step}}}, \frac{\text{step} \times \text{warmup\_steps}^{-1.5}}{\sqrt{\text{warmup\_steps}}}\right) \times c
\]

where \(d_{\text{Embedding}}\) is the embedding dimension of the model, \( \text{warmup\_steps} \) is the number of warm-up steps and \( c \) is an additional multiplier coefficient. 

The hyperparameters during the training process are shown in Table \ref{tab:hyper}. We acknowledge that due to limited computational resources, the hyperparameters are not optimized, especially for the largest PMT-550M model, where significant convergence had not been observed. There is still room for improvement in model performance on the same dataset and architecture.

\begin{table}[!htbp]
  \caption{Hyperparameters for training. The format "*/*" denotes the hyperparameters for the next location prediction task and mask imputation task, respectively.}
  \centering
  \begin{tabular}{@{}l l c c c c c c@{}}
    \toprule
    & Model & Label smoothing & Dropout & Warmup steps & \(c\) & Batch Size & Epoch\\
    \midrule
    \multirow{3}{*}{\rotatebox[origin=c]{90}{BOS}} 
    & PMT-1.6M    & 0.1 & 0.1 & 1,000/1,000 & 1/1 & 32/16 & 20/20 \\
    & PMT-21M     & 0.1 & 0.1 & 1,000/1,000 & 1/1 & 32/16 &  40/20 \\
    & PMT-550M    & 0.1 & 0.3 & 5,000/5,000 & 1/1 & 32/16 &  40/20 \\
    \midrule
    \multirow{3}{*}{\rotatebox[origin=c]{90}{LA}} 
    & PMT-1.6M    & 0.1 & 0.1 & 1,000/1,000 & 1/1 & 32/16 & 20/10 \\
    & PMT-21M     & 0.1 & 0.1 & 1,000/1,000 & 1/1 & 32/16 & 20/10 \\
    & PMT-550M    & 0.1 & 0.3 & 20,000/2,000 & 1/0.05 & 32/16 & 20/10 \\
    \midrule
    \multirow{3}{*}{\rotatebox[origin=c]{90}{NYC}} 
    & PMT-1.6M    & 0.1 & 0.1 & 1,000/1,000 & 1/0.01 & 32/16 & 20/10 \\
    & PMT-21M     & 0.1 & 0.1 & 1,000/1,000 & 1/1& 32/16 & 20/10 \\
    & PMT-550M    & 0.1 & 0.3 & 40,000/2,000 & 1/0.05 & 32/16 & 15/4\\
    \bottomrule
  \end{tabular}
  \label{tab:hyper}
\end{table}

\section{Evaluation}
\subsection{Baselines}
In this section, we describe the baseline models used for various tasks. We detail their hyperparameters in Table \ref{tab:baseline}.

\textbf{Frequency-based Model (FBM).} This method follows a simple logic, where users are most likely to appear where they most frequently visit. In its implementation, we calculate the locations where users most frequently appear during daytime and nighttime, and provide the top k predictions based on frequency. The frequency of users appearing at different locations is computed based on data from the entire study period.

\textbf{Long Short-Term Memory (LSTM).} LSTM \cite{hochreiter1997long} is a typical recurrent neural network for sequence prediction problems. The unidirectional nature of LSTM models limits the utilization of future information. Therefore, for imputation tasks, we used a bidirectional LSTM (BiLSTM) for comparison.

\textbf{DeepMove.} DeepMove \cite{feng2018deepmove} is a model designed for mobility prediction. The model leverages historical and recent user mobility patterns using the combination of gated recurrent units (GRUs) and an attention module. To extend the capability of this model in imputation tasks, we employed bidirectional GRU layers. This modified model is referred as BiDeepMove.

\begin{table}[!htbp]
  \caption{Hyperparameters of baseline models.}
  \centering
  \begin{tabular}{@{}lccccccc@{}}
    \toprule
    \multirow{2}{*}{Model} & \multicolumn{3}{c}{Embedding Size} & \multirow{2}{*}{Hidden Size} & \multirow{2}{*}{RNN Layers}  & \multirow{2}{*}{Total Parameters}\\
    \cmidrule(lr){2-4}
          & Location & Time & User & & & \\
    \midrule
    LSTM & 128 & -   & - & 256  & 2             & 7M\\
    BiLSTM & 128 & -   & - & 256  & 2            & 12M\\
    DeepMove  & 512 & 512 & 64 & 1024 & 3  & 60M\\
    BiDeepMove  & 512 & 512 & 64 & 1024 & 3  & 130M\\
    \bottomrule
    
  \end{tabular}
  \label{tab:baseline}
\end{table}

\subsection{Metrics for generation evaluation}

\textbf{Modified n-gram Precision.} Given the generated trajectory sequence \(\hat{X} = \{\hat{x}_1, \hat{x}_2, \ldots, \hat{x}_T\}\) and the reference sequence \(X = \{x_1, x_2, \ldots, x_T\}\). The modified n-gram precision function $p_n(\hat{X}; X)$ is defined as:

\begin{equation*}
p_n(\hat{X}; X) = \frac{\sum_{s\in G_n(X)} \min(C(s, \hat{X}), C(s, X))}{\sum_{s\in G_n(X)} C(s, X)}
\end{equation*}

Where $G_n(X)$ is the set of n-grams of length $n$ in sequence $X$. Any n-grams containing unknown locations will not be included, e.g., for \(X = \{a, b, [\text{nan}], b, c, c, c\}\), $G_2(X) = \{ab, bc, cc, cc\}$, $G_3(X) = \{bcc, ccc\}$. $C(s, X)$ is count of n-gram $s$ in $G_n(X)$.

\textbf{Aggregated Level Metrics.} Given the generated trajectory sequence $\hat{X} = \{\hat{x}_1, \hat{x}_2, \ldots, \hat{x}_T\}$ and the corresponding reference sequence $X = \{x_1, x_2, \ldots, x_T\}$, we only consider $T'$ for aggregation, where $X_t \neq [\text{nan}]$ for all $t \in T'$. This ensures that the sum of aggregations is identical between the generated and reference trajectories.

While root mean square error (\textbf{RMSE}) is computed for all areas, mean absolute percentage error (\textbf{MAPE}) is calculated for the top 25\% of CBGs with the highest population. This strategy mitigates potential distortions in \textbf{MAPE} caused by extremely low ground truth values.

\section{Supplementary Results}

\begin{table}[!htbp]
  \caption{Performance of the imputation task. The format "*/*" denotes the cross entropy (no label smoothing) for the training set (included in pretraining) and the testing set (excluded in pretraining), respectively.}
  \label{tab:imputation}
  \centering
  \begin{tabular}{l@{\hspace{15pt}}lccccc}
    \toprule
     & \multirow{2}{*}{Model} & \multicolumn{5}{c}{Removal rate}  \\
    \cmidrule(r){3-7}
    & & 50\% & 60\% & 70\% & 80\% & 90\% \\
    \midrule
    \multirow{3}{*}{\rotatebox[origin=c]{90}{BOS}} 
    & PMT-1.6M    & 0.72/0.75 & 0.76/0.79 & 0.82/0.85 & 0.91/0.95 & 1.12/1.16  \\
    & PMT-21M     & 0.62/0.64 & 0.66/0.68 & 0.72/0.75 & 0.81/0.84 & 1.00/1.03  \\
    & PMT-550M    & \textbf{0.57}/\textbf{0.59} & \textbf{0.61}/\textbf{0.64} & \textbf{0.67}/\textbf{0.70} & \textbf{0.77}/\textbf{0.80} & \textbf{0.93}/\textbf{0.96}  \\
    \midrule
    \multirow{3}{*}{\rotatebox[origin=c]{90}{LA}} 
    & PMT-1.6M    & 1.15/1.20 & 1.20/1.25 & 1.27/1.32 & 1.40/1.44 & 1.62/1.67  \\
    & PMT-21M     & 0.72/0.73 & 0.77/0.78 & 0.84/0.86 & 0.96/0.97 & 1.17/1.18  \\
    & PMT-550M    & \textbf{0.70}/\textbf{0.71} & \textbf{0.75}/\textbf{0.76} & \textbf{0.82}/\textbf{0.84} & \textbf{0.92}/\textbf{0.94} & \textbf{1.10}/\textbf{1.13}  \\
    \midrule
    \multirow{3}{*}{\rotatebox[origin=c]{90}{NYC}} 
    & PMT-1.6M    & 1.70/1.73 & 1.73/1.76 & 1.78/1.81 & 1.86/1.89 & 2.02/2.06  \\
    & PMT-21M     & 0.75/0.76 & 0.80/0.81 & 0.87/0.88 & 0.99/1.00 & 1.20/1.22  \\
    & PMT-550M    & \textbf{0.66}/\textbf{0.67} & \textbf{0.71}/\textbf{0.72} & \textbf{0.77}/\textbf{0.79} & \textbf{0.87}/\textbf{0.89} & \textbf{1.05}/\textbf{1.07}  \\
    \bottomrule
  \end{tabular}
\end{table}

